# Decoding Urban-health Nexus: Interpretable Machine Learning Illuminates Cancer Prevalence based on Intertwined City Features

Chenyue Liu[1*], Ali Mostafavi[2]

[1]Ph.D. Student, Urban Resilience.AI Lab, Zachry Department of Civil and Environmental Engineering, Texas A&M University, College Station, United States; e-mail: liuchenyue@tamu.edu

[2]Associate Professor, Urban Resilience.AI Lab Zachry Department of Civil and Environmental Engineering, Texas A&M University, College Station, United States; e-mail: amostafavi@civil.tamu.edu

## Abstract

This study investigates the interplay among social demographics, built environment characteristics, and environmental hazard exposure features in determining community-level cancer prevalence. Utilizing data from five Metropolitan Statistical Areas in the United States — Chicago, Dallas, Houston, Los Angeles, and New York—the study implemented an XGBoost machine learning model to predict the extent of cancer prevalence and evaluate the importance of different features. Our model demonstrates reliable performance, with results indicating that age, minority status, and population density are among the most influential factors in cancer prevalence. We further explore urban development and design strategies that could mitigate cancer prevalence, focusing on green space, developed areas, and total emissions. Through a series of experimental evaluations based on causal inference, the results show that increasing green space and reducing developed areas and total emissions could alleviate cancer prevalence. The study and findings contribute to a better understanding of the interplay among urban features and community health and also show the value of interpretable machine learning models for integrated urban design to promote public health. The findings also provide actionable insights for urban planning and design, emphasizing the need for a multifaceted approach to addressing urban health disparities through integrated urban design strategies.

# 1. Introduction

Public health outcomes in cities arise from the interplay of complex and nonlinear interactions among urban features. Yet, the existing approaches for formulating public health policies in cities focus primarily on a limited number of features using statistical methods that assume linear relationships between features and health outcomes. This limitation has hindered integrated urban design strategies to reduce disease prevalence in cities by failing to consider the salutary effects of improving the built environment and environmental hazard characteristics in different areas of cities. To address this gap, in this study, we examine the extent to which urban features related to the built environment, socio demographic characteristics, and environmental hazards and their non-linear interactions shape the prevalence of cancer using interpretable machine learning models.

The incidence of cancer and its associated morbidity and mortality rates pose significant public health challenges worldwide. The American Cancer Society estimates that there will be more than 1.9 million new cancer cases diagnosed and more than 600,000 cancer deaths in the United States in 2023 alone [1]. While these statistics are indeed alarming, they also underscore the urgent need for uncovering the factors contributing to cancer prevalence to inform preventative strategies. Despite the substantial advancements in understanding the genetic [2] [3] [4] and lifestyle factors [5] [6] associated with cancer, less attention has been given to the potential role of the built environment and environmental hazard exposures. Recent research, however, is beginning to highlight the significant effects that social demographics [7], urban development, and environmental hazards factors [8] can have on public health outcomes, including cancer.

From an urban health perspective, features related to the built environment, socio-demographic characteristics, and environmental hazards could shape the prevalence of cancer in different areas of a city. In particular, the built environment, characterized by human-made physical structures and infrastructures, can have direct and indirect impacts on cancer outcomes. Urban design features such as green spaces have proven to provide health benefits [9], including stress reduction and opportunities for physical activity, which may in turn reduce cancer risk. On the other hand, urban areas with high levels of development [10] are often associated with increased exposure to environmental hazards such as air pollution [11] and heat islands [12], which have been linked to a variety of adverse health effects, including cancer.

Further, the role of environmental hazard exposures, such as particulate matter (PM2.5) and other air pollutants, in cancer incidence and prevalence is increasingly recognized. Long-term exposure to PM2.5 has been associated with increased mortality from lung cancer and has also been linked to other cancers, such as breast and bladder cancer. The exposure to such environmental hazards occurs in close interaction with the built environment characteristics of urban areas. Thus, to achieve a comprehensive perspective, it is essential to capture and evaluate the complex and non-linear interactions among various urban features that shape cancer prevalence in cities. The existing statistical approaches, however, fail to capture non-linear interactions among features, and most of the extant studies focus on distinct types of urban features, such as environmental hazards in isolation when evaluating cancer prevalence.

Given the complex interplay between urban features in shaping the prevalence of cancer in cities, there is a growing need for comprehensive, data-driven research approaches that can capture and quantify these interactions to predict and explain cancer prevalence in communities. Machine-learning (ML) methods, with their ability to model complex, non-linear relationships and interactions between variables, hold significant promise in this regard.

Recognizing this important gap, this study is geared towards developing and applying an interpretable machine-learning model to predict and explain cancer prevalence in communities based on social demographics, urban development, and environmental hazard exposures. Specifically, the study uses the XGBoost algorithm, a powerful and efficient ML technique, along with the Synthetic Minority Oversampling Technique (SMOTE) for addressing class imbalance, the SHapley Additive exPlanations (SHAP) method for interpreting the model, and causal inference for deriving policy recommendations. This study focuses on five Metropolitan Statistical Areas (MSAs): Chicago, Dallas, Houston, Los Angeles, and New York. Accordingly, we aim to answer three important research questions:

1. To what extent can a machine learning model that captures the interactions between social demographics, built environment, and environmental hazard exposures reliably predict cancer prevalence in communities?
2. What is the level of importance for different features in explaining the variability of cancer prevalence in communities?
3. What urban development and design features or strategies could alleviate the prevalence of cancer in communities?

The findings from this study will contribute to our understanding of the urban features and their interplay in shaping cancer prevalence in urban communities and provide valuable insights for urban planning and public health interventions aimed at reducing cancer risk. The study also serves as a model for how machine learning techniques can be used to enhance our understanding of complex public health issues and other urban phenomena and inform integrated urban design strategies for promoting urban sustainability and health.

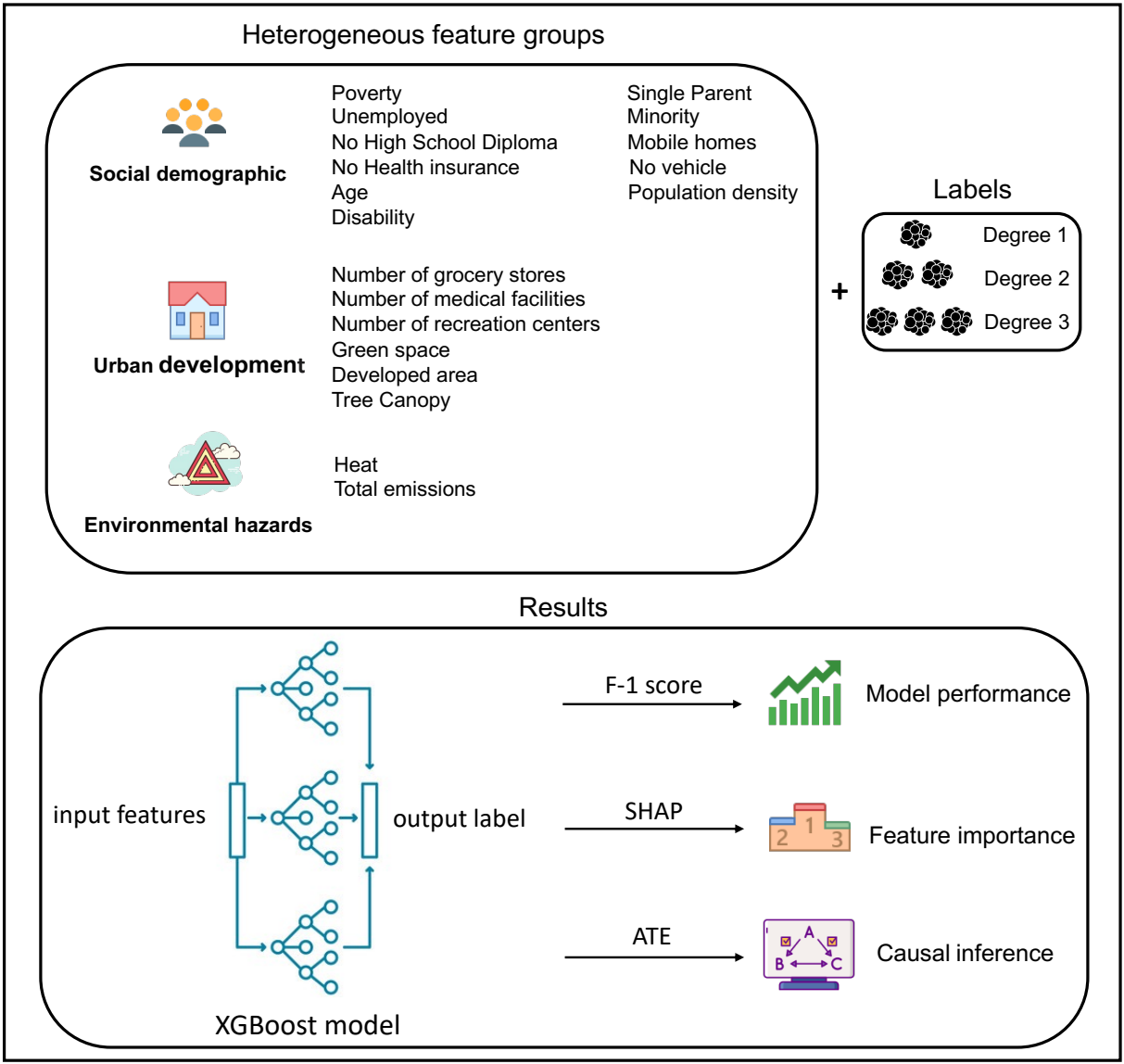

*Figure 1.* **Overview of research framework:** Social demographic characteristics, urban development, and environmental hazards data are used to predict the degree of cancer. After giving the above features as an input, the XGBoost model which is verified as the best model will give the predicted label for each census tract among 5 MSAs. Then the model performance is evaluated by the F-1 score. Feature importance and causal inference are determined by SHAP value and ATE value which can be used to identify significant factors and guide policy recommendations.

## 2. Data Description

In this study, we focus on five metropolitan statistical areas (MSAs) within the United States: Chicago-Naperville-Elgin, Dallas-Fort Worth-Arlington, Houston-The Woodlands-Sugar Land, Los Angeles-Long Beach-Anaheim, and New York-Newark-Jersey City. To ensure the representativeness of the study areas, the selection process involved criteria such as population size, geographic distribution, and dataset availability. Specifically, the chosen areas were required to exhibit a sufficiently large population size to facilitate the examination of cancer prevalence variations across different neighborhoods. Moreover, the selection also prioritized areas situated in diverse regions across the United States, allowing for the inclusion of multiple geographic variations. Last but not least, the study also considered the availability of data in the selected areas, as data availability is a critical factor in feature calculation in this study. By implementing these selection criteria, the study aimed to ensure the robustness and generalizability of the findings, thereby enhancing their potential applicability to a broader population. Table 1 summarizes the features examined in this study. For each city, we calculated these features at the census tract level and then trained machine learning models to predict the extent of cancer prevalence based on these features.

*Table1.* Description of Variables.

| Theme | Variable abbreviation | Description | Data type |
|---|---|---|---|
| Socio-demographics | Poverty | % people whose income is below the 150% federal poverty level | continuous |
| | Unemployed | % unemployed people | |
| | NHD (No High school Diploma) | % people with no high school diploma (age 25+) | |
| | NHI (No health insurance) | % uninsured in the total civilian noninstitutionalized population | |
| | Age (Age greater than 65) | % people aged 65 and older | |
| | Disability | % people with a disability | |
| | SP(Single parent) | % single-parent households with children under 18 | |
| | Minority | % minorities | |
| | MH(Mobile homes) | % people who own mobile homes | |
| | NV(No vehicle) | % households with no vehicle available | |
| | PD(Population density) | The number of individuals per square kilometer | |
| Urban development | #GS | The Number of the grocery stores | |
| | #MF | The Number of the medical facilities | |
| | #RC | The Number of the recreation centers | |
| | GS (Green space) | % percentage of people living within 0.5 miles of a park | |
| | DA (Developed area) | % land used for development in low intensity, medium intensity, and high intensity | |
| | TC (Tree canopy) | % land covered by deciduous forests, evergreen forests, and mixed forests | |
| Environmental hazards | Heat | the number of extreme heat days occurring between May and September of year 2019 | |
| | TE (Total emissions) | the annual average levels of PM2.5 for the entire year of 2019 | |
| Health outcome | Cancer | degree of prevalence of cancer | categorical |

## 2.1. Socio-demographic features

The social vulnerability index (SVI), was developed by the Centers for Disease Control and Prevention (CDC) [13]. In recent decades, many studies are investigating the relationships between social vulnerability-related metrics and cancer. Tran et al. gave a deep review of studies that applied the SVI to investigate outcomes in different types of cancer patients [14]. Ganatra et al. studied the impact of social vulnerability on mortality attributed to comorbid cancer and found that age-adjusted mortality rates for comorbid cancer have higher mortality in counties with greater social vulnerability [15]. The impact of US county-level SVI on breast and colon cancer screening rates is highlighted by Mehtal et al., which emphasizes the need for more effective intervention strategies and resource allocation to improve SDOH for the most vulnerable citizens in the country [16]. In this study, we examine features related to poverty, unemployed, no high school diploma, no health insurance, age greater than 65, disability, minority, mobile home, and no vehicle. More details about the features are described below:

*Poverty:* The federal poverty line thresholds established for several federal health coverage policies [17]. The percentage of population in poverty in each census tract was defined as the percentage of people whose income is below the 150% federal poverty level. *Unemployed:* Percentage of unemployed people. *No high school diploma:* Percentage of people with no high school diploma (age 25+). *No health insurance:* Percentage of uninsured in the total civilian noninstitutionalized population. *Age greater than 65:* Percentage of persons aged 65 and older. *Disability:* Percentage of the civilian noninstitutionalized population with a disability. *Single parent:* Percentage of single-parent households with children under 18. *Minority:* Percentage of minorities (Hispanic or Latino, Black and African-American, American-Indian and Alaska Native, Asian). *Mobile home*: Percentage of people who own mobile homes. *No vehicle:* Percentage of households with no vehicle available. *Population density:* A few decades ago, the population density was proven by several studies to be a very important factor for cancer occurrence and mortality. Among both genders in Taiwan, Yang, and Hsieh observed a significant increase in mortality rates for cancers of the lung, pancreas, and kidney with more urbanization [18]. A statistically significant linear trend was observed by Nasca et al. among males and females for cancers of the buccal cavity and pharynx, esophagus, bronchus and lung, stomach, and colon, indicating an increase in incidence with increasing population density [19]. In this study, the population density feature is expressed as the number of individuals per square kilometer.

## 2.2. Urban development data and features

The built environment characteristics that shape urban development patterns also can influence public health outcomes in cities. In this study, we examined urban development features related to the distribution of points-of-interest (POI), as well as green space, tree canopy, and developed area features. These features capture characteristics of urban development that could shape public health outcomes. The number of grocery stores and medical facilities and the number of recreation centers data were retrieved from Safegraph [20]. SafeGraph serves as a reliable and authoritative source of information for the physical world, offering an accurate and current account of the geographic locations of places, including basic information as well as categorization. The data furnished by SafeGraph is of superior quality and is highly dependable. The information we used here is the 4-digit North American Industry Classification System (NAICS) code and the geographical location describing each POI. Federal statistical agencies have adopted NAICS [21] as their standard method of categorizing business establishments.

By using the NAICS code information, we can filter out codes other than those of grocery stores, medical facilities, and recreation centers. We use the geographic location information to count the number of POIs at the census-tract level. The NAICS code and its corresponding category we used are shown in the supplementary information. *Number of the grocery store:* The number of grocery stores in a specific census tract. *Number of medical facilities:* The number of medical facilities in a specific census tract. *Number of recreation centers:* The number of recreation centers in a specific census tract.

The green space, developed area, and tree canopy data are from National Environmental Public Health Tracking Network [22] which is operated by the Centers for Disease Control and Prevention. This network encompasses comprehensive data and information regarding environmental factors, hazards, health outcomes, and population health.

*Green space:* The percentage of people living within 0.5 miles of a park. The data leveraged the area-proportion technique to ascertain the number of individuals residing within a half-mile radius of publicly accessible parks at the census tract level. To this end, the data computed the proportion of each census tract falling within the 0.5-mile radius of a park and multiplied this fraction by the corresponding population to yield an estimate of the number of people residing in the relevant segment of the census tract. Subsequently, we obtained the percentage of individuals residing in proximity to parks by dividing the estimated population by the total number of individuals residing within the census tract. *Developed area:* Percentage of land used for low-intensity, medium-intensity, and high-intensity development. *Tree canopy:* Percentage of land covered by deciduous forests, evergreen forests, and mixed forests.

## 2.3. Environmental hazards data and features

Environmental hazards such as heat and pollution are shown to be closely linked with cancer prevalence. We examined the features related to environmental hazards using the datasets explained here. *Heat:* The metric The count of extreme heat days occurring between May and September. The raw data for Heat is captured through a gridded format with an approximate resolution of 14 x 14 km, as defined by the North American Land Data Assimilation System (NLDAS) [23]. In order to aggregate the raw data to the census tract level, the United States Census Block Group centroids were used to attribute individual NLDAS grid cells through a containment relationship. Specifically, the maximum daily temperature for each block group was computed by identifying the highest hourly value for each day. Subsequently, population-weighted averages were used to calculate the maximum hourly value for the census tract level. The threshold for extreme temperature was defined based on the 90th percentile of temperature observed from May through September 2020. The number of extreme heat days at the census tract level was determined by comparing each day's maximum daily temperature with this threshold.

*Total emissions:* The toxicity data are retrieved by Community Multiscale Air Quality(CMAQ) modeling system developed by Environmental Protection Agency(EPA) [24]. The CMAQ modeling system integrates crucial physical and chemical processes related to the dissemination and alteration of air contaminants across different ranges. It covers the entire continental United States using 12-km by 12-km grid cells. The model's temporal adaptability allows for yearly to multi-year assessments of pollutant climatology and movement from concentrated origins through simulations in time periods of weeks or months. In this study, air pollution data are in the annual average PM2.5 level in 2019.

## 2.4. Health outcome

Cancer prevalence, the focus of this analysis, obtained from the Centers for Disease Control and Prevention [13], captures the prevalence of cancer (excluding skin cancer) in percentage form. This study employs the equal-frequency binning method to partition cancer prevalence into three distinct clusters. The data are initially sorted in ascending order by the percentage of cancer, resulting in clusters with the same width or range of values. Specifically, Census tracts with cancer category equal to one denotes low prevalence, while census tracts in category three have relatively high cancer prevalence. A pertinent issue arises in that the above approach yields imbalanced samples within each cluster. To address this, we employ the Synthetic Minority Over-sampling Technique algorithm, as explained further in the methodology section.

## 3. Method

We examined different machine learning techniques to identify the best model for addressing the research questions in this study. The overview of the methods evaluated in this study are presented below.

## 3.1. K-Nearest Neighbors (KNN)

The fundamental concept of the K-Nearest Neighbors (KNN) algorithm [25] involves utilizing a training dataset to classify new input instances. Specifically, when giving a new input, the model identifies K-nearest neighbors by putting the new instance into the training dataset. These K-nearest neighbors among the new instance serve as a reference point for determining the classification of it, with the majority class among the K-nearest neighbors ultimately dictating the classification outcome. The algorithm's efficacy is contingent on the suitability of the training dataset, as well as the appropriateness of the K value selected. The value of K is automatically determined and is optimal in terms of the F-1 score.

## 3.2. Decision tree

The objective of utilizing a Decision Tree algorithm [26] is to construct a tree model that can accurately predict the class or value of the target variable by learning straightforward decision rules deduced from the provided training data. The construction of the decision tree commences from the root and, at each iteration, endeavors to identify features and formulate a condition that partitions all the classes within the dataset to the highest level of purity possible. The degree of purity is quantified using either the Gini impurity or Shannon information gain metrics. This iterative process continues until each sample within the training dataset is situated within its own cluster, thereby obtaining the optimal F-1 score.

### 3.3. Random forest

The Random Forest algorithm [27] is a tree-based model that employs an ensemble of decision trees. Each tree is constructed using a random subset of features and data points randomly sampled from the training set. Once each decision tree is trained, the algorithm utilizes a majority vote strategy to determine the predicted cluster of a new data point by aggregating the predictions of all the trees within the forest.

### 3.4. XGBoost

The implementation of XGBoost [28] entails the iterative addition of trees and feature splits to grow the model. Each new tree learns a function that fits the residual predicted from the previous iteration. The model parameters are optimized in each leaf node for every tree, with control over the maximum number of trees and the depth of each tree. To predict the score of a sample, the algorithm locates the corresponding leaf node for that sample in each tree based on its features, with each leaf node corresponding to a score. Subsequently, aggregating the scores from each tree provides the final prediction for the new data point.

### 3.5. Performance Evaluation Standards

When evaluating the performance of machine learning models in traditional binary classification problems, several standards can be used, including precision, recall, accuracy, and F-1 score. These metrics are derived from the confusion matrix, which comprises four categories: True Negative (TN), False Positive (FP), False Negative (FN), and True Positive (TP). True Negative refers to examples that are correctly labeled as negative, False Positive refers to negative examples that are incorrectly labeled as positive, True Positive corresponds to positive examples that are correctly labeled as positive, and False Negative are positive examples that are incorrectly labeled as negative. Precision is a widely used metric in binary classification and is calculated as the number of true positives divided by the sum of false positives and true positives. Precision measures the proportion of positive predictions that are correctly classified. A high precision score indicates that the model is correctly predicting the positive class, while a low precision score indicates that the model is misclassifying positive examples as negative. The precision can be calculated using the following formula:

$$\text{Precision} = \frac{\text{TP}}{\text{TP} + \text{FP}}$$

Recall, also known as sensitivity or true positive rate, is another metric used to evaluate the performance of machine learning models in binary classification problems. Recall measures the proportion of actual positive examples that are correctly identified by the model. It is calculated as the number of true positives divided by the sum of false negatives and true positives:

$$\text{Recall} = \frac{TP}{TP + FN}$$

Accuracy is a widely used metric in machine learning to evaluate the overall performance of binary classification models. It measures the percentage of correctly classified examples, both positive and negative. Accuracy is calculated as the sum of true positives and true negatives divided by the total number of examples:

$$\text{Accuracy} = \frac{TP + TN}{(TP + TN + FP + FN)}$$

The F-1 score, also known as the F-measure or F-score, is a harmonic mean of precision and recall. It is a widely used metric in binary classification problems when both precision and recall are equally important. The F-1 score is calculated as the weighted average of precision and recall:

$$F - 1 \text{ score} = 2 \times \frac{\text{Precision} \times \text{Recall}}{(\text{Precision} + \text{Recall})}$$

As our model is a multi-classification problem, we need to modify the evaluation metrics for assessing the model's performance. In this case, we can define the positive outcome as the predicted label that matches the true label for each example. For each cluster, we can calculate the four categories of the confusion matrix: True Negative (TN), False Positive (FP), False Negative (FN), and True Positive (TP). In this paper, we use the macro-average F-1 score to evaluate the model's overall performance, as we have three clusters. The macro-average F-1 score calculates the F-1 score for each cluster and then takes the average across all clusters. This approach provides equal weight to each cluster, regardless of the number of examples in each cluster. The macro-average precision, recall, and F-1 score for each cluster can be calculated as follows, where 1, 2, and 3 represent different clusters:

$$\text{Macro Precision} = \frac{P_1 + P_2 + P_3}{3}$$

$$\text{Macro Recall} = \frac{R_1 + R_2 + R_3}{3}$$

$$\text{Macro F} - 1 \text{ Score} = \frac{\text{Macro Precision} \times \text{Micro Recall}}{(\text{Macro Precision} + \text{Macro Recall})}$$

### 3.6.  Synthetic Minority Oversampling Technique

A pertinent issue in clustering the census tracts based on cancer prevalence is imbalanced samples within each cluster. To address this, we employ the Synthetic Minority Over-sampling Technique algorithm, as explained further in the methodology section. The fundamental idea behind SMOTE algorithm [29] is grounded in the principle of analyzing and simulating minority class samples, subsequently augmenting artificially simulated new samples to the dataset to alleviate the class imbalance in the original data. The K-Nearest Neighbor technique serves as the bedrock for the algorithm's simulation process, wherein the following steps are employed for generating new samples via simulation: First, the sampling nearest neighbor algorithm is employed to compute the K nearest neighbors of each minority class sample. Secondly, N samples are randomly selected from the K nearest neighbors for random linear interpolation. Thirdly, new minority class samples are constructed using the interpolated values. Finally, the new samples are merged with the original data to generate a new training set.

### 3.7. SHapley Additive exPlanations

The objective of our analysis is not to predict cancer prevalence across census tracts, but rather to examine the extent to which various urban features and their non-linear interactions shape cancer prevalence. Hence, the interpretability of machine learning models plays a key role in this study. Interpretable machine learning is aimed at comprehending the rationale behind machine learning model predictions. The SHAP method has emerged as a prominent approach to address this challenge and has been previously validated for its interpretability performance [30]. The Shapley Additive exPlanations (SHAP) method was introduced by Lundberg and Lee [19], who adapted a concept from game theory [31] to interpret the predictions of complex training processes in machine learning models based on the SHAP values. These values determine the importance of features in the model, with larger SHAP values indicating greater significance. Following a comparison of classification models, the model with the highest Macro F-1 score was selected. The SHAP method was subsequently applied to calculate the SHAP values, which were used to determine the relative importance of features.

### 3.8. Causal inference

Interpretable machine learning methods such as SHAP inform about mainly feature importance. To formulate urban design strategies, however, it is essential to conduct causal experiments on important features to examine the extent to which changes on the features improve or exacerbate cancer prevalence in cities. To determine the causal connection between two variables, the causal inference [32] method assesses the effect of a treatment variable on an outcome variable while taking into account any confounding factors. The purpose of causal inference is to identify cause-and-effect relationships, which are essential for making informed decisions, and for constructing efficient interventions and regulations. It is hard, however, to figure out the genuine causal effect when a third factor has an influence on both the treatment and the result, a phenomenon known as confounding, which makes causal inference complex. To address confounding issue, we use a controlled variable technique that will hold all the confounding variables constant. Then by calculating the average treatment effect (ATE), we can reveal the causal between the outcome and manipulated variable. ATE is defined as:

$$\text{ATE} = E[Y(W = 1)] - E[Y(W = 0)]$$

where $Y(W = 1)$ and $Y(W = 0)$ are the potential treated outcome and control group outcome of the whole population respectively.

After learning the causal effect between some determinants and the percentage of cancer and by training a reliable model, we can predict the overall percentage of cancer in each MSA by changing the value of the determinant. Comparing this value (treated value) with the original value (controlled value) before change, we can quantify the ATE.

## 4. Results

### 4.1. Evaluating Cancer Prevalence Models

Each census tract is characterized by a set of features that are introduced in the data description section. The dependent variable is the cancer prevalence cluster to which the census tract belongs. The dependent variable takes on values of 1, 2, or 3; and higher values of the cancer cluster indicate a greater prevalence of cancer. In the training process, 70% of the census tracts are used as the training set, and 30% as a test set. To prevent overfitting, we also apply five-fold cross-validation. To determine the optimal hyperparameters for each model, we evaluate different options and calculate the macro F-1 score for each one. The hyperparameter that yields the highest macro F-1 score is selected as the best choice. Here, each MSA will have an XGBoost model with a different set of hyperparameters.

To address Research Question 1, we evaluated the performance of different models for predicting the prevalence of cancer (Table 2). While the random forest model demonstrated comparable performance to the XGBoost model for certain metropolitan statistical areas, we ultimately chose the XGBoost model as the primary model for this project due to its overall better performance.

*Table 2.* Macro F-1 score of models in prediction of the prevalence of cancer.

|  | KNN | Decision tree | Random Forest | XGBoost |
|---|---|---|---|---|
| Chicago | 0.66 | 0.70 | 0.78 | 0.79 |
| Dallas | 0.55 | 0.57 | 0.64 | 0.66 |
| Houston | 0.57 | 0.62 | 0.63 | 0.66 |
| Los Angeles | 0.70 | 0.82 | 0.79 | 0.83 |
| New York | 0.68 | 0.80 | 0.83 | 0.84 |

During the training process of the XGBoost model, six hyperparameters were set to control the behavior of the algorithm: learning rate, number of estimators, subsample, colsample_bytree, max depth, alpha, and gamma.

The learning rate hyperparameter governs the speed at which the XGBoost model adapts to the problem by controlling the magnitude of each update to the model's parameters. The number of estimators hyperparameter determines the number of decision trees that the model builds during training. The subsample hyperparameter controls the fraction of observations, or rows, that are randomly sampled for each decision tree. The colsample_bytree hyperparameter governs the fraction of features, or columns, that are randomly sampled for each decision tree. The max depth hyperparameter restricts the maximum depth of each decision tree to avoid overfitting.

Finally, the regularization parameter alpha and gamma control the trade-off between the model's complexity and its ability to generalize to new data. Specifically, alpha is the L1 regularization term that adds a penalty proportional to the absolute value of the coefficients, while gamma is the minimum loss reduction required to make a split in a decision tree node. By setting appropriate values for these hyperparameters, the XGBoost model can avoid overfitting and achieve better performance on unseen data.

In addition to assessing model performance, we implemented a confusion matrix a tool to evaluate classification models. Figure 2 displays the confusion matrix, with the x-axis representing the predicted class and the y-axis representing the actual class for each census tract in each metropolitan statistical area. The results indicate that the XGBoost model has the ability to accurately predict between three classes. Specifically, within the Dallas MSA, 68% of census tracts in class 1 and 81% of census tracts in class 2 are correctly predicted. Although not performing as well as the other classes, the model is still able to correctly predict half of the census tracts in class 3. These outcomes instill confidence in the ability of the model to address the research questions pertaining to the downward trend.

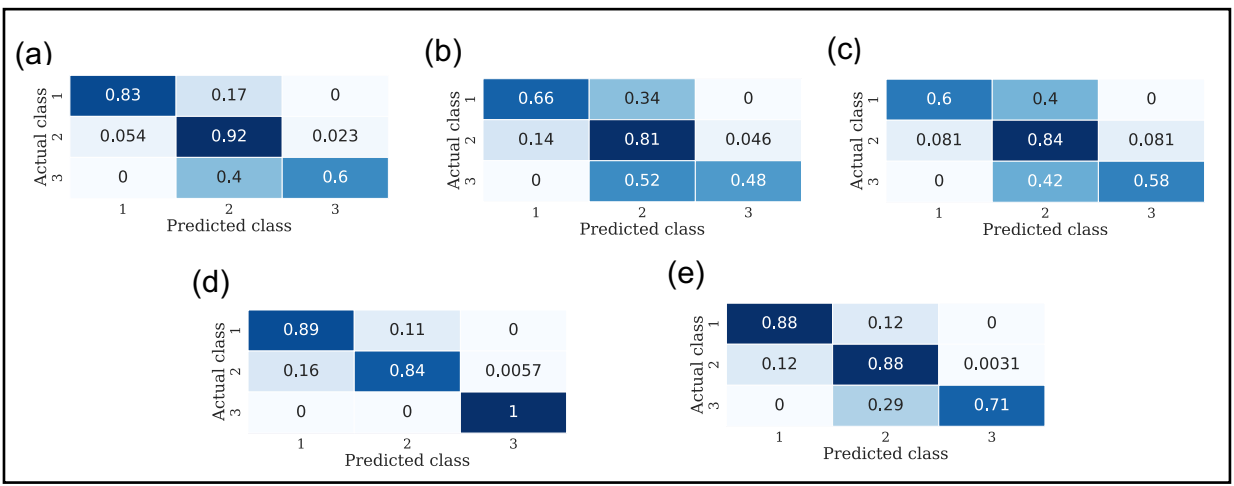

*Figure 2.* **Confusion matrix for different degrees of cancer in five different MSAs by XGBoost model**: (a) Chicago-Naperville-Elgin. (b) Dallas-Fort Worth-Arlington. (c) Houston-The Woodlands-Sugar Land. (d) Los Angeles-Long Beach-Anaheim. (e) New York-Newark-Jersey City. (1, 2, and 3 stand for clusters indicating increasing degrees of cancer prevalence.)

## 4.2. SHAP Analysis of Important Features

Table 3 presents the distribution of census tracts based on their primary determinants. Intuitively, the feature of age greater than 65 ranks as the most prominent factor influencing the degree of cancer in 57.98% of the MSAs. This feature is followed by minority status with 23.1%, population density with 3.19%, no vehicles with 3.12%, and no high school diploma with 2.63%. These results prompted us to investigate the distribution of these important features across the five MSAs. To explore this, we generated a geographical plot (Figure 4) to visualize the potential spatial patterns.

*Table 3.* Counties have the same factor with the highest importance.

| Most important features | Number of census tracts | Share of census tracts (%) |
|:---:|:---:|:---:|
| Age | 7781 | 57.98 |
| Minority | 3100 | 23.10 |
| PD | 428 | 3.19 |
| NV | 419 | 3.12 |
| NHD | 353 | 2.63 |
| heat | 228 | 1.70 |
| TE | 225 | 1.68 |
| Disability | 191 | 1.42 |
| #NMF | 147 | 1.10 |
| #NG | 111 | 0.83 |
| NHI | 91 | 0.68 |
| SP | 85 | 0.63 |
| Poverty | 80 | 0.60 |
| #NR | 60 | 0.45 |
| Unemployed | 41 | 0.31 |
| DA | 39 | 0.29 |
| MH | 20 | 0.15 |
| GS | 13 | 0.10 |
| TC | 7 | 0.05 |

As discussed previously, Age greater than 65 dominates most census tracts for the Houston, Los Angeles, and New York MSAs, as well as the central area of the Dallas MSA. In the Chicago MSA, total emission (PM2.5) is concentrated in the middle eastern part. Population density is distributed to the outlier of the Dallas MSA, which is due to the sparsely populated areas. Minority is distributed in the eastern part of the Los Angeles MSA. No clear patterns can be discerned for the Houston and New York MSAs.

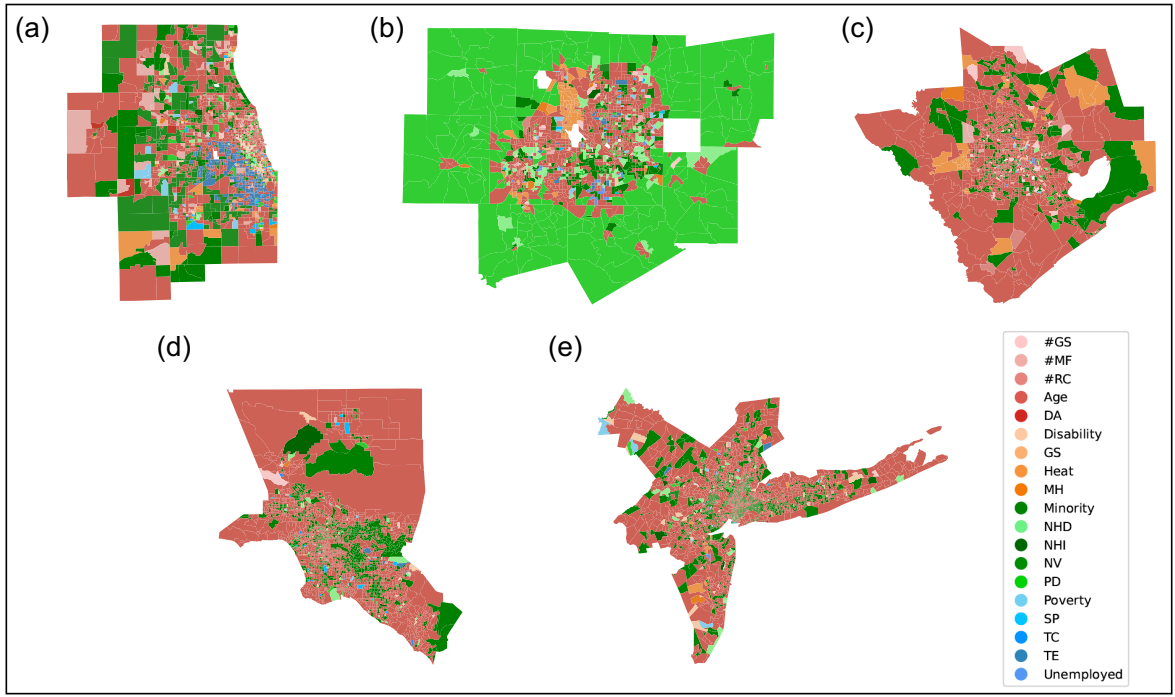

*Figure 3.* Geographical Distribution of the Most Important Feature Across Five Key Metropolitan Statistical Areas. The figure illustrates the spatial distribution of the feature with the highest importance across the following MSAs: (a) Chicago-Naperville-Elgin, (b) Dallas-Fort Worth-Arlington, (c) Houston-The Woodlands-Sugar Land, (d) Los Angeles-Long Beach-Anaheim, and (e) New York-Newark-Jersey City.

In this study, we applied the XGBoost model and the SHAP method to analyze the degree of cancer across 5 Metropolitan Statistical Areas (MSAs), using a large set of 19 variables. By training the XGBoost model, we were able to obtain the best hyperparameters, resulting in the highest Macro F-1 score and optimal model performance. We then applied the SHAP method to rank the feature importance, as shown in Figure 4 (a)(b)(c)(d)(e), which illustrates the importance of each variable across different MSAs.

Our results indicate that the percentage of people who are older than 65 is the most important feature across all five MSAs, which is consistent with our prior knowledge that older individuals are more susceptible to cancer. In Chicago, social vulnerability metrics and environmental hazard features (total emissions and heat) are among the top 10 most important variables related to the degree of cancer. In Dallas, social vulnerability metrics, urban development features, and environmental hazard features (heat) are among the top 10 most important features. In Houston, social vulnerability metrics, urban development features, and environmental hazard features (heat) are also among the top 10 most important features. In Los Angeles, social vulnerability metrics and environmental hazard features (heat and total emissions) are among the top 10 most important features. In New York, social vulnerability metrics, urban development features, and environmental hazard features (heat) are among the top 10 most important features.

Notably, the number of medical facilities and the number of recreation centers are only present in the Dallas MSA's top 10 important features. Heat appears in all MSAs, while total emissions are only present in Chicago and Los Angeles. These findings suggest that areas with high heat exposure has a greater prevalence of cancer than areas with higher total emissions. Overall, the results highlights the importance of considering various social, environmental, and urban development features when analyzing the degree of cancer across different MSAs. Upon scrutiny of results in Figure 4 (f), it is evident that the Chicago MSA with the highest impact of heat in terms of the degree of cancer. Conversely, the remaining metropolitan statistical areas (MSAs) exhibit comparable effects of heat. With regard to total emissions, the MSA with the greatest impact of total emissions on the prevalence of cancer is also Chicago, followed by Los Angeles, Houston, Dallas, and New York.

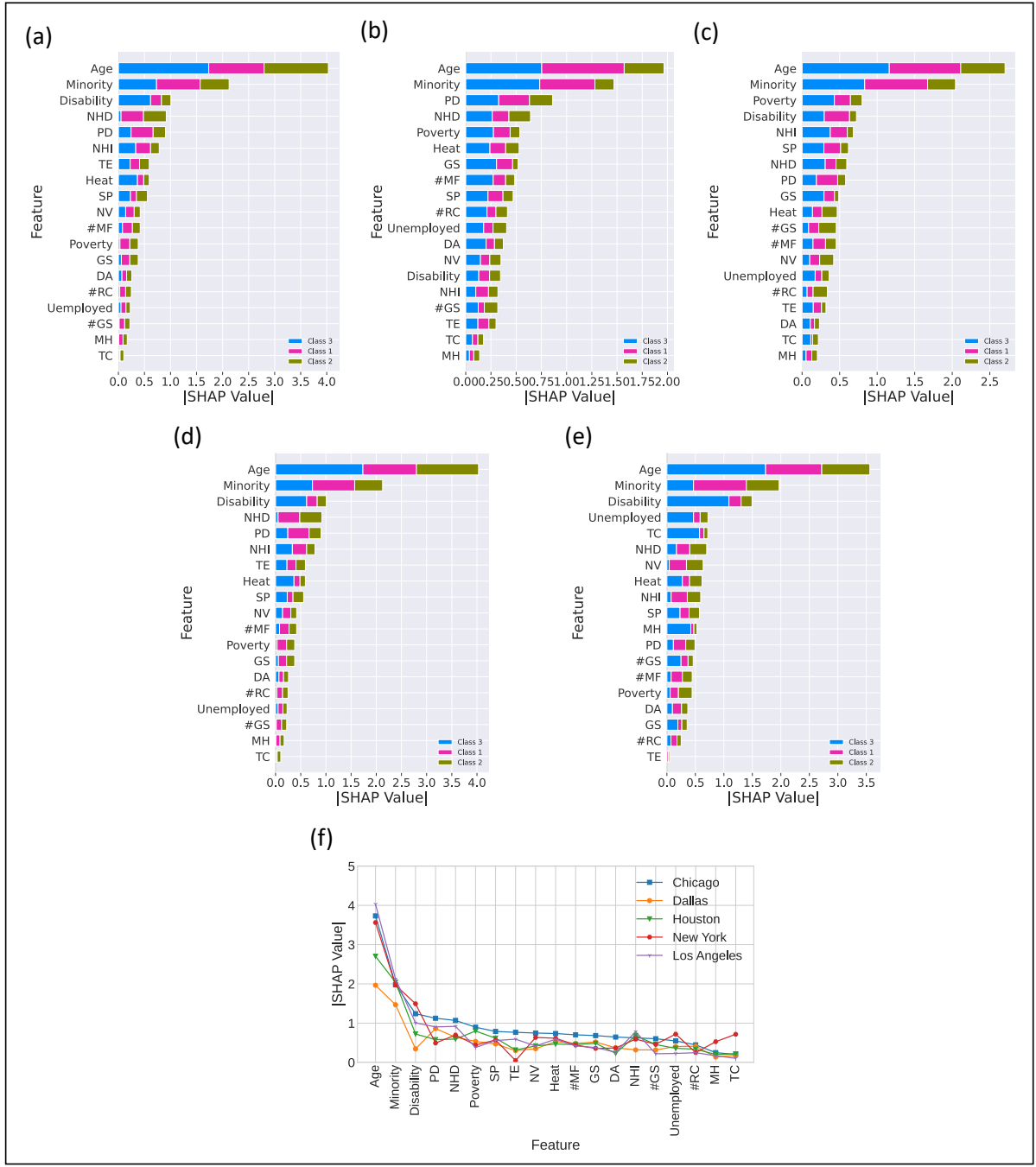

*Figure 4.* Comparative Importance of Features Across Different Metropolitan Statistical Areas (MSAs) Measured by the Absolute SHAP Values. The SHAP values are presented on the x-axis, with larger values indicating higher feature importance. The features are labeled on the y-axis. The represented MSAs are (a) Chicago-Naperville-Elgin, (b) Dallas-Fort Worth-Arlington, (c) Houston-The Woodlands-Sugar Land, (d) Los Angeles-Long Beach-Anaheim, (e) New York-Newark-Jersey City, and (f) an aggregation of the five MSAs.

## 4.3. Causal inference experiments

While feature importance analysis informs about the significance of different features in explaining variability in the prevalence of cancer across census tracts in the MSAs, it does not offer the necessary causal relationship insights to inform urban design and policy strategies. For this purpose, in the next step, we identify urban design strategies that can reduce the prevalence of cancer in high-risk areas. Accordingly, we have chosen to examine important features related to urban development and environmental hazards that can be influenced by policy changes. These features include green space, developed area, and total emission, all of which are significant predictors of the prevalence of cancer, despite some of them not being ranked among the top 10 important features in the SHAP analysis except that the SHAP value for total emissions in New York MSA is zero. Based on the results presented in the SHAP summary plot, the aggregated SHAP values of these factors do not differ significantly from those of the tenth most important feature in the five metropolitan statistical areas analyzed.

To examine the causal relationship between certain features and cancer prevalence, relevant scientific evidence must be considered. Several studies have investigated the association between these features and cancer incidence, providing valuable insights into their potential effects. Demoury et al. suggest that exposure to natural environments may have a favorable impact on cancer risk. [9]. Bowler et al. did a meta-analysis that provided some evidence of a positive benefit of a walk or run in a natural environment in comparison to a synthetic environment [33]. Furthermore, the assumption that residential green space plays a role in decreasing the risk of certain cancer types is supported by the research of Pope III et al [11]. Greenberg showed that urban areas have higher cancer mortality and cancer incidence rates than rural areas [10].

By considering these studies, we can gain insights into the potential causal relationship between these features and cancer prevalence.

Following the establishment of relationships between the identified features and the degree of cancer prevalence, we conducted a series of experiments to explore potential interventions. In particular, we focused on green space and employed a two-step approach. First, we calculated the mean value of the degree of cancer ($E[Y(W = 0)]$) associated with relatively high prevalence (cluster 2 and cluster 3) based on the predicted cluster obtained from the original XGBoost model trained in Section 4.1. Second, we increased the green space value by 25% by adding more parks located within 0.5 mile of residential areas. To ensure that other factors remained constant, we employed the same hyperparameters used in the original model and calculated the mean value of the degree of cancer ($E[Y(W = 1)]$) associated with relatively high severity (cluster two and cluster three). We then calculated the Average Treatment Effect, which is the difference between $E[Y(W = 1)]$ and $E[Y(W = 0)]$.

*Table 4.* Average Treatment Effect results.

| MSA | ATE | | |
|---|---|---|---|
| | Green space | Developed areas | Total emissions |
| Chicago | -0.019 | -0.015 | -0.019 |
| Dallas | -0.027 | -0.03 | -0.032 |
| Houston | -0.005 | -0.008 | -0.01 |
| Los Angeles | -0.036 | -0.035 | -0.036 |
| New York | -0.039 | -0.04 | NA |

Accordingly, we conducted two experiments. In the first experiment, then we increased the developed green areas by 25% and computed the ATE for each MSA. In the second experiment, we decreased the total emissions by 25% and computed the ATE for each MSA, except for the New York MSA, where the total emission feature had a zero SHAP value. The findings of these experiments offer causal insights into the potential impact of urban design interventions on the degree of cancer and can inform the development of effective strategies to mitigate the adverse effects of urban features on cancer prevalence. The results, show that all the ATE values for green space, developed area, and total emissions were negative among the five MSAs (Table 4). This result suggests that increasing the green space, decreasing the developed area, and decreasing the total emissions in the census tracts could alleviate the degree of cancer in these areas. Moreover, the results reveal that for green space and developed areas, the most affected MSA was New York, while for total emissions, the most affected MSA was Los Angeles. These findings underscore the different effectiveness extent of different strategies in different cities and highlight the importance of implementing city-specific urban design interventions to alleviate cancer prevalence in high-risk areas of in the studied MSAs. After examining the city-level strategies, we focused on the census tract level and examine the extent to which the characteristics of the census tracts compare with improved cancer prevalence with all census tracts within the MSA. First, we used maps to visualize the distribution of areas that exhibited changes from high to low degrees of cancer. Our findings suggest that the patterns of these changed areas are consistent throughout the MSA. A detailed depiction of the distribution of these improved areas can be found in the figure in the Supplementary information.

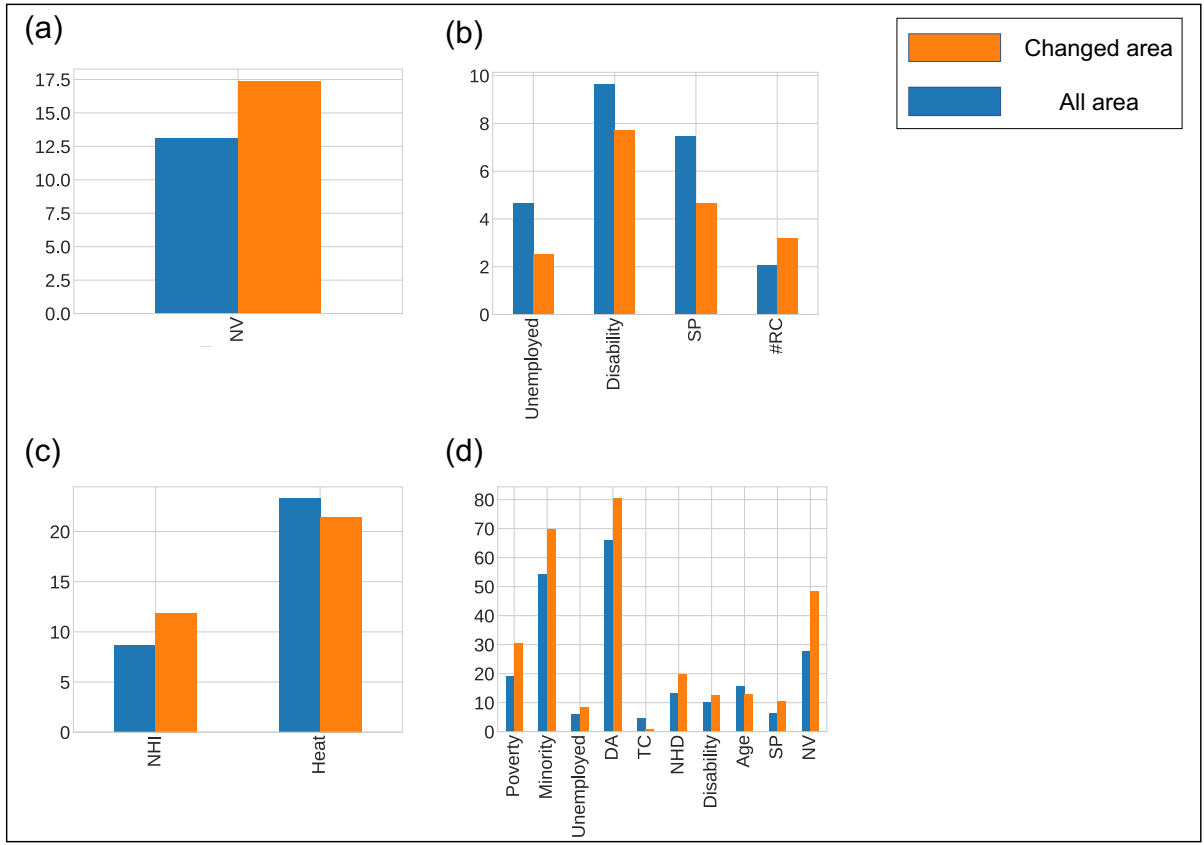

*Figure 5.* **Differences of characteristics of the changed area and all areas for five MSAs.** (a) Chicago-Naperville-Elgin MSA. (b) Dallas-Fort Worth-Arlington MSA. (c) Los Angeles-Long Beach-Anaheim MSA. (d) New York-Newark-Jersey City MSA.

In the next step, we analyzed the characteristics of changed areas versus all areas in an MSA (Figure 5). The term "all areas" refers to the mean value of features across all census tracts in the MSA, while "changed areas" refers to census tracts that exhibited a decrease in the degree of cancer following the modification of the green space, developed area, and total emissions values. A t-test was performed on the means of the two groups; only the features with a p-value smaller than 0.05 were considered to exhibit a significant difference between them.

The Houston MSA exhibited no significant differences between the means of the two groups, and hence the results were not displayed in Figure 5. Conversely, the changed areas in the Chicago MSA had a significantly higher percentage of people with no vehicle. In the Los Angeles MSA, the changed areas had a higher percentage of people with no health insurance and encounter fewer heat days in summer, indicating greater vulnerability to health and economic risks. Similarly, in the New York MSA, the changed areas were identified as more vulnerable and had large populations with disabilities. These findings show that the identified urban interventions would particularly reduce cancer prevalence in areas where vulnerable populations reside, and ultimately would reduce urban health inequality.

# 5. Closing Remarks

Urban health outcomes are multifaceted phenomena, born out of intricate and non-linear interplay among diverse urban characteristics. Existing methodologies that shape urban design strategies and public health policies in cities, however, are largely concentrated on a limited array of features, utilizing statistical methods underpinned by the assumption of linear relationships between said features and health outcomes. This restricted approach posed obstacles to the development of comprehensive urban design strategies aimed at curbing disease prevalence, by enhancing the built environment and adjusting environmental hazard attributes in varying city areas. Departing from limitations of the existing methods, this study delves into the degree to which aspects of the built urban development, socio-demographic features, and environmental hazards—along with their non-linear interconnections—influence cancer prevalence, deploying interpretable machine learning models for the purpose.

The selected XGBoost model showcased promising performance, making it a reliable tool for explaining variations in cancer prevalence across census tracts based on their heterogeneous features. Its ability to accurately predict different cancer prevalence classes in the studied MSAs underscores its utility in health-related urban planning. We identified different determinants of cancer prevalence in the communities studied, with age greater than 65 and minority being the most significant across all five MSAs. This is consistent with established knowledge about cancer susceptibility, reaffirming the model's validity. Notably, environmental hazard features, particularly heat, emerged as significant predictors in all MSAs, while total emissions were only significant in Chicago and Los Angeles. This result suggests different environmental factors could play varying roles in cancer prevalence in conjunction with other urban development and socio-demographic features. The study also examined the importance of green space, developed area, and total emissions to evaluate effective urban design interventions. By manipulating these factors through the use of causal inference, we were able to estimate their impact on cancer prevalence. Increasing green space and decreasing developed area and total emissions all led to a decrease in cancer degree, as indicated by the negative average treatment effect values. These findings suggest that urban design interventions targeting these elements could play a crucial role in mitigating cancer prevalence in urban areas. Further, the geographical analysis of areas that exhibited changes in cancer degrees revealed notable patterns. These areas were often more vulnerable and less desirable as residential locations, highlighting the role of socio-economic factors in health outcomes. This suggests that interventions should focus not only on environmental and urban design aspects, but also take into account social vulnerability and inequality.

This study, therefore, illuminates the potential for data-driven integrated urban design in improving public health outcomes. It emphasizes the importance of considering heterogeneous urban features, ranging from social demographics to environmental hazards, and their complex interactions in urban design, planning and policymaking. Moreover, it underscores the need for targeted, context-specific interventions, given the variability in the importance of different factors across different MSAs.

Future research could adopt the machine learning and causal inference approaches in this study in examining other urban sustainability problems. Different urban health and sustainability phenomena are emergent properties in cities arising as a result of complex interactions among various urban features. Hence, the approach presented in this paper can inform future studies to better decode complex urban sustainability problems using machine learning, departing from the standard reductionist and linear methods. Such machine learning methods could hold the key to integrated urban design strategies to promote urban sustainability and health by evaluating the ways in which the interplay among various urban features shape different outcomes.

## Acknowledgement

This material is based in part upon work supported by Texas A&M University X-Grant 699. The authors also would like to acknowledge Spectus and SafeGraph for providing mobility and population activity data. Any opinions, findings, conclusions or recommendations expressed in this material are those of the authors and do not necessarily reflect the views of the Texas A&M University, Spectus and SafeGraph.

## Data availability

The data that support the findings of this study are available from SafeGraph and Spectus, but restrictions apply to the availability of these data, which were used under license for the current study. The data can be accessed upon request submitted to the data provides. Other data we use in this study are all publicly available.

## Code availability

The code that supports the findings of this study is available from the corresponding author upon request.